\pdfoutput=1
\documentclass[11pt]{article}

\usepackage[review]{EMNLP2022}

\usepackage{times}
\usepackage{latexsym}

\usepackage[T1]{fontenc}
\usepackage[utf8]{inputenc}

\usepackage{microtype}

\usepackage{inconsolata}

\usepackage{graphicx}

\usepackage{hyperref}

\usepackage{color}

\title{BBT-Fin: Comprehensive Construction of Chinese Financial Domain Pre-trained Language Model, Corpus and Benchmark}

\author{Dakuan Lu\textsuperscript{\rm 1},
\textbf{Hengkui Wu}\textsuperscript{\rm 3}\thanks{\enspace Corresponding author.},
Jiaqing Liang\textsuperscript{\rm 2},
Yipei Xu\textsuperscript{\rm 1},
Qianyu He\textsuperscript{\rm 1}, \\
\textbf{Yipeng Geng}\textsuperscript{\rm 3},
\textbf{Mengkun Han}\textsuperscript{\rm 3},
\textbf{Yingsi Xin}\textsuperscript{\rm 3},
\textbf{Yanghua Xiao}\textsuperscript{\rm 1}\footnotemark[1]
\\
            \textsuperscript{\rm 1}Shanghai Key Laboratory of Data Science, School of Computer Science, Fudan University\\
    \textsuperscript{\rm 2}School of Data Science, Fudan University\\
    \textsuperscript{\rm 3}SuperSymmetry Technologies\\
    \{ludakuan1234, l.j.q.light, xuyipei000, abbey4799\}@gmail.com, \\
    \{ypgeng, mkhan, ysxin, hkwu\}@ssymmetry.com, shawyh@fudan.edu.cn
    }
\begin{document}
\maketitle
\begin{abstract}
To advance Chinese financial natural language processing (NLP), we introduce BBT-FinT5, a new Chinese financial pre-training language model based on the T5 model. To support this effort, we have built BBT-FinCorpus, a large-scale financial corpus with approximately 300GB of raw text from four different sources. In general domain NLP, comprehensive benchmarks like GLUE and SuperGLUE have driven significant advancements in language model pre-training by enabling head-to-head comparisons among models. Drawing inspiration from these benchmarks, we propose BBT-CFLEB, a Chinese Financial Language understanding and generation Evaluation Benchmark, which includes six datasets covering both understanding and generation tasks. Our aim is to facilitate research in the development of NLP within the Chinese financial domain.
Our model, corpus and benchmark are released at \url{https://github.com/ssymmetry/BBT-FinCUGE-Applications}. Our work belongs to the Big Bang Transformer (BBT), a large-scale pre-trained language model project.
\end{abstract}

\section{Introduction}
Pre-trained language models(PLMs), such as BERT \cite{devlin2018bert} and T5 \cite{2019t5}, have led to great performance boosts across many NLP tasks. Despite the excellent performance of pre-trained language models (PLMs) on a large number of NLP tasks, their performance is often affected when applied to domain-specific texts that exhibit significant differences from general text in terms of word usage, syntax, and writing style \cite{gururangan2020don,gu2021domain}. To address this issue, \citet{gururangan2020don} proposed that continuing to pre-train a general PLM on target domain corpora and task-relevant texts can effectively improve its performance on domain-specific tasks, while \citet{gu2021domain} further suggested that pre-training domain-specific PLMs from scratch with a sufficiently large corpus can achieve even better domain-specific performance. Inspired by these studies, domain-specific pre-trained language models have emerged in some domains, such as BioBERT~\cite{peng-etal-2019-transfer} and PubMedBERT~\cite{gu2021domain} in the biomedicine field, which have been utilized for practical tasks like entity and relation extraction.

We collect all existing NLP competition tasks and academic datasets related to finance on the Chinese internet and summarized them in Table~\ref{tab:fin-datasets}, revealing a growing demand for NLP capabilities in finance, particularly in information extraction and sentiment analysis. To meet these demands and improve the overall level of Chinese financial NLP, several companies have already developed and released Chinese financial pre-trained language models, such as FinBERT~\cite{FinBERT} and Mengzi-BERT-base-fin~\cite{zhang2021mengzi}. However, these models are based on the BERT-base model, have a single architecture type, and a parameter count (around 110 million) that is outdated and unable to meet the increasing demand for NLP capabilities in this field. Therefore, we propose FinT5, the largest Chinese financial pre-trained language model to date, based on the advanced T5 architecture, with 220 million parameters for the base version and 1 billion for the large version.

Furthermore, NLP tasks in the financial industry focus primarily on information extraction, requiring models with high entity knowledge understanding and memorization capabilities. Although studies have shown that pre-trained PLMs on large-scale corpora already have some entity knowledge understanding and memorization capabilities, there are still some shortcomings. To address this issue, many studies have used knowledge-enhanced pre-training methods to improve PLMs' understanding and memorization of entity knowledge. However, these methods mostly target BERT-like models and lack strategies designed for T5 models. To improve T5's performance on financial NLP tasks, we propose a concise knowledge-enhanced pre-training method based on the T5 model's text-to-text paradigm.

In addition, another challenge faced by Chinese financial NLP is the lack of corpus. The scale and diversity of corpora play an essential role in language model pre-training \cite{xu2020clue,2019t5,gao2020pile}. However, existing Chinese financial corpora are small in scale, poor in diversity and not open, as can be shown in Table \ref{tab:typicalplm-corpus}. To solve this problem, we first need to determine the text types that a qualified Chinese financial corpus needs to cover. To this end, we first collected almost all existing Chinese financial NLP tasks and summarized their text sources, as shown in the Table~\ref{tab:fin-datasets}. According to the source distribution of these tasks, we have determined the range of text types we need to collect. As a result, we collect and release a large-scale Chinese financial corpus named BBT-FinCorpus with about 300 GB raw text, which consists of five different sources to enhance its diversity covering most text sources of Chinese financial NLP tasks. 

The widespread use of benchmark evaluations is a key driving force that has greatly improved and rapidly iterated PLMs. These evaluations use a single score to assess model performance across multiple tasks, enabling direct and comprehensive comparisons between pre-trained language models. Existing English PLMs use the general benchmark evaluations GLUE \cite{wang2018glue} and SuperGLUE \cite{wang2019superglue}, while the general benchmark evaluation for Chinese PLMs is CLUE \cite{xu2020clue}. Almost all PLMs participate in these evaluations to compare their performance with other models. However, there is no publicly available benchmark for Chinese financial NLP, which makes it difficult to compare existing pre-trained language models on different task sets and hinders the rapid improvement of PLM performance in the Chinese financial domain.

To address this issue and promote research in the financial domain, we propose CFLEB, the \textbf{C}hinese \textbf{F}inancial \textbf{L}anguage Understanding and Generation \textbf{E}valuation \textbf{B}enchmark, consisting of six datasets covering language understanding and generation tasks. These datasets encompass a diverse range of text genres, dataset sizes, and levels of difficulty, and more importantly, emphasize challenges that arise in real-world scenarios.

Our contributions are summarized as follows:
\begin{itemize}
    \item We introduce BBT-FinT5, a state-of-the-art financial Chinese PLM with large-scale parameters and knowledge-enhanced pre-training.
    \item We provide BBT-FinCorpus, a comprehensive and diverse financial Chinese corpus.
    \item We propose BBT-CFLEB, a benchmark for evaluating Chinese language understanding and generation in the financial domain.
\end{itemize}

\section{Related Work}
% 1.金融领域模型-语料库：FinBERT(https://arxiv.org/abs/1908.10063，英文，使用基于关键词过滤出的金融新闻作为语料库，在金融情绪分析任务上评测），
% FinBERT(https://arxiv.org/abs/2006.08097，英文，使用金融通讯语料库训练，在金融情绪分析任务上评测）
% FinBERT(https://github.com/valuesimplex/FinBERT，中文，在30亿token的新闻/研报/公告/百科上训练，在短讯分类，情绪分析，NER任务上验证）
% Mengzi-BERT-base-fin(https://arxiv.org/abs/2110.06696, 中文，在20GB新闻/公告/研报上训练，在新闻分类，情绪分析任务上验证）
\subsection{Domain-specific PLMs and Corpora}
PLMs have achieved state-of-the-art performance in many NLP tasks ~\cite{devlin2018bert,2019t5,liu2019roberta}. However, when applied to domain-specific tasks, models pre-trained on general corpora often produce unsatisfactory results due to the difference in word distribution from general to specific domains ~\cite{gururangan2020don, gu2021domain}. To better adapt a language model to a target domain, pre-training on the corpus of the target domain is proposed ~\cite{gururangan2020don}. For domains with abundant unlabeled text, such as biomedicine, pre-training from scratch results in substantial gains over continual pre-training of general-domain language models ~\cite{gu2021domain}. Consequently, many domain-specific PLMs have been proposed and pre-trained on their respective corpora.

In the field of financial NLP, domain-specific pre-trained language models (PLMs) have demonstrated their superiority over general-domain PLMs. For instance, ~\citet{araci2019finbert} and ~\citet{yang2020finbert} pre-trained BERT on English finance news and communications, respectively, and outperformed competitive baselines on financial sentiment analysis tasks. In the context of Chinese financial NLP, ~\citet{FinBERT} pre-trained BERT on Chinese financial news, analysis reports, company announcements, and encyclopedias, and evaluated it on news classification, sentiment analysis, and named entity recognition tasks. Furthermore, ~\citet{zhang2021mengzi} pre-trained the Chinese PLM Mengzi on a 20GB financial corpus and demonstrated its effectiveness on multiple downstream tasks.

Table~\ref{tab:typicalplm-corpus} summarizes the characteristics of typical PLMs and their corpora in the financial domain. It can be observed that both the scale of our model and corpus exceed existing works.

\begin{table*}[!htb]
\centering
\begin{tabular}{p{5cm} l l p{6cm}}
\hline
\textbf{PLM} & \textbf{Size} & \textbf{Corpus Size} & \textbf{Corpus Sources}\\
\hline
FinBERT~\cite{araci2019finbert}
& 110M
& 29M words
& News filtered by financial keywords 
\\
FinBERT~\cite{yang2020finbert}
& 110M
& 4.9B tokens
& Corporate Reports, Earnings Call Transcripts, Analyst Reports
\\
FinBERT~\cite{FinBERT}
& 110M
& 3B tokens
& News, Analyse reports, Company announcements and Encyclopedias
\\
Mengzi-BERT-base-fin~\cite{zhang2021mengzi}
& 110M
& 20GB file
& News, Analyse reports, Company announcements
\\
BBT-FinT5 (ours)
& 220M, 1B
& 80B tokens
& Corporate Reports, Analyst Reports, Social media and Financial News
\\
\hline
\end{tabular}
\caption{Typical financial PLMs and their corpora.}
\label{tab:typicalplm-corpus}
\end{table*}

\subsection{Knowledge Enhanced Pre-training}
Although PLMs can acquire rich linguistic knowledge from pretraining on large-scale corpora, many studies have shown that PLMs still have shortcomings in entity knowledge understanding and memory, as the distribution of entity knowledge in unfiltered corpora is sparse and long-tailed~\cite{yang2021survey}. Therefore, PLMs can benefit from knowledge-enhanced pretraining methods that strengthen entity knowledge understanding and memory.

For example, Ernie~\cite{sun2019ernie} is designed to learn language representation enhanced by knowledge masking strategies, which includes entity-level masking and phrase-level masking. The disadvantage of this approach is that it can only help the model better learn existing entity knowledge from the corpus, without addressing the issues of sparse and long-tailed distribution of entity knowledge in the corpus.

Ernie 3.0, introduced by \citet{sun2021ernie}, incorporates the universal knowledge-text prediction (UKTP) task. This task involves a pair of triples from a knowledge graph and their corresponding sentences from an encyclopedia, where either the relation in the triple or the words in the sentence are randomly masked. In order to predict the relation in the triple, the model must identify the head and tail entities mentioned in the sentence, and determine the semantic relationship between them.

The limitation of this approach is that it only masks the relation in the triple and not the entities, which can hinder the learning of entity representations. Moreover, distant supervision has a certain amount of noise, which means that the relation in the triple may not necessarily appear in the sentence~\cite{smirnova2018relation}. Therefore, only masking the relation and predicting it can have a strong negative impact on the model.
Although the above methods have made some progress, they are all designed for the BERT-like model. 

To our knowledge, there is currently a gap in knowledge enhancement pre-training methods available for T5-like models.

% 2.评测基准：目前没有金融领域的评测基准，现有的金融领域预训练模型使用了多种不同的任务进行垂直（相对于通用模型）评测，无法进行横向（相对于其他金融模型）评测
% 其他领域的评测基准：生物医学领域有BLUE（https://aclanthology.org/W19-5006/ 首个生物医学领域评测基准），
% BLURB（https://dl.acm.org/doi/full/10.1145/3458754 改良版BLUE，删除了两个不相干的临床医学任务，同时提出了对整体平均分数的改进方案）

\subsection{Domain-specific NLP Benchmarks}
Various domain-specific NLP benchmarks have been proposed to compare the ability of different methods in modeling text from specific domains in a fair manner.
The BLUE benchmark~\cite{peng2019transfer} evaluates the ability of models in biomedical text mining through five tasks.
The BLURB benchmark~\cite{gu2021domain} further focuses on clinical domains by removing two unrelated tasks and includes a wider range of biomedical applications.
Despite these efforts, a comprehensive set of benchmark tasks for training, evaluating, and analyzing financial PLMs is still largely unexplored.
Currently, the FLUE~\cite{shah2022flue} is the only benchmark for the financial domain, consisting of five tasks specifically designed for English financial text.
However, we are the first to construct a comprehensive set of benchmarks for Chinese financial text, covering a range of language understanding and generation tasks that differ from previous works.

% \subsection{How to compute a summary score}
% To compute a summary score for a benchmark with multiple tasks, the simplest way is to report the average score among all tasks.
% However, this may place undue emphasis on simpler tasks such as NER for which there are many existing datasets. Therefore, \citet{10.1145/3458754} group the datasets by their task types, compute the average score for each task type, and report the macro average among the task types. 

% Furthermore, using the average of model scores on dataset neglects characteristics of different datasets and metrics. For instance, the numerical results of BLEU\cite{papineni2002bleu} are typically small as compared with F1 scores. As a result, with score averaging, the overall score can be dominated by metrics with larger scale. To address the issue, \citet{papineni2002bleu} normalize the scores based on scores of representative standard baseline models, i.e., T5-Small\cite{2019t5}, so as to eliminate the influence of disturbing factors such as different metrics. Specifically, the normalized score on dataset is given by p/b, where p and b are dataset performance of model under evaluation and the standard baseline model respectively. Normalization on standard model performance essentially gives a ratio to different metrics and datasets, therefore making the overall score more reasonable.

\section{The Corpus: BBT-FinCorpus}
\label{sec:fincorpus}
We build FinCorpus, the biggest corpus of Chinese financial domain to get a superior pre-trained language model. Section \ref{sec:corpus-scale} covers how we decided on the corpus contents. We collected, refined and sorted the corpus to finally obtain the FinCorpus, as elaborated in Section \ref{sec:fincorpus-desc}.

\begin{table*}[!htb]
\centering
\resizebox{2\columnwidth}{!}{
\begin{tabular}{p{7cm} p{4cm} l p{2cm}}
\hline
\textbf{Dataset} & \textbf{Text Source} & \textbf{Open State} & \textbf{Practicality}\\
\hline
DuEE-fin~\cite{han2022duee}
& Financial news, Company announcement
& Yes
& High 
\\
FinRE~\cite{li-etal-2019-chinese}
& Financial news
& Yes
& High
\\
Announcement information extraction~\cite{ieaalc}
& Company announcement
& Yes
& High
\\
Discovery of new entities in Internet finance~\cite{df-internet}
& Social media
& Unspecified
& Low
\\
Announcement information extraction~\cite{bd-public}
& Company announcement
& Unspecified
& High
\\
Construction of financial knowledge graph~\cite{bd-kg}
& Analyse report
& Unspecified
& Medium
\\
Event causality extraction~\cite{bd-cau}
& Financial news
& Unspecified
& Low
\\
Financial NL2SQL~\cite{bd-sql}
& Data query sentence
& Unspecified
& Medium
\\
Few-shot event extraction~\cite{bd-few}
& Financial news
& Unspecified
& Medium
\\
Few-shot event extraction~\cite{bd-trans}
& Financial news
& Unspecified
& Medium
\\
FinNL (ours)
& Financial news
& Yes
& High
\\
FinNA (ours)
& Financial news
& Yes
& High
\\
FinFE (ours)
& Social media
& Yes
& High
\\
FinNSP (ours)
& Social media
& Yes
& High
\\
\hline
\end{tabular}
}
\caption{Chinese financial datasets we collected, with their open source status and practicality scores}
\label{tab:fin-datasets}
\end{table*}

\subsection{Coverage Confirmation of the Corpus}
\label{sec:corpus-scale}
We believe that, since the purpose of domain pre-training is to help models better understand domain texts and perform domain tasks more effectively, it is essential to observe the text distribution of domain tasks to determine the coverage of the corpus. The domain corpus should cover the text sources of domain tasks as much as possible to enhance the model's understanding of the tasks. To this end, we first collected almost all Chinese financial NLP task datasets available on the Chinese internet in recent years, including several datasets used in this study, and their text sources, as shown in Table \ref{tab:fin-datasets}.

It can be seen that the text sources of these financial NLP datasets are mainly concentrated in financial news, company announcements, research reports, and social media. For financial news, we chose the largest financial news websites on the Chinese Internet for crawling, namely Sina Finance~\footnote{\url{https://finance.sina.com.cn/}}, Tencent Finance~\footnote{\url{https://new.qq.com/ch/finance/}}, Phoenix Finance~\footnote{\url{https://finance.ifeng.com/}}, 36Kr~\footnote{\url{https://36kr.com/}}and Huxiu~\footnote{\url{https://www.huxiu.com/}}. For company announcements and research reports, we chose Eastmoney~\footnote{\url{https://www.eastmoney.com/}} for crawling. For social media, we chose the two largest financial social media platforms on the Chinese Internet, Guba~\footnote{\url{https://guba.eastmoney.com/}} and Xueqiu~\footnote{\url{https://xueqiu.com/}}, for crawling.

\subsection{Crawling and Filtering of the Corpus}
We used a proxy-based distributed crawler to crawl public web pages. We filtered the web pages using a series of rules~\cite{2019t5,YUAN202165}.

\subsection{Description of the Corpus}
\label{sec:fincorpus-desc}
After crawling, cleaning, and processing, we obtained the FinCorpus, a large-scale Chinese financial domain corpus that contains four types of language materials:
\begin{itemize}
\item \textbf{Corporate announcements.} \quad These are the announcements released by all listed companies in China over the past twenty years. The original data is in PDF format, with a total size of about 2TB. Using a PDF parser, we converted the PDF files into text files, resulting in a total size of 105GB.
\item \textbf{Research reports.} \quad These are research reports issued by investment institutions such as securities firms and investment banks on macroeconomic issues, sectors, industries, and individual stocks, analyzing the current status and future development trends of the research object. The original data is in PDF format, with a total size of about 1TB. After conversion, the total size of the resulting text files is about 11GB.
\item \textbf{Financial news.} \quad These are the financial news articles from the past five years crawled from websites including Sina Finance, Tencent Finance, Phoenix Finance, 36Kr, and Huxiu. After cleaning, the total size of the resulting text files is about 20GB.
\item \textbf{Social media.} \quad These are the posts from all stockholders and bloggers published on stock bar and Xueqiu website over the past twenty years. After cleaning, the total size of the resulting text is about 120GB.
\end{itemize}
The corpus from the above five sources basically covers all types of texts in the common Chinese financial NLP.

\begin{figure*}[!htb]
    \centering
    \includegraphics[width=2\columnwidth]{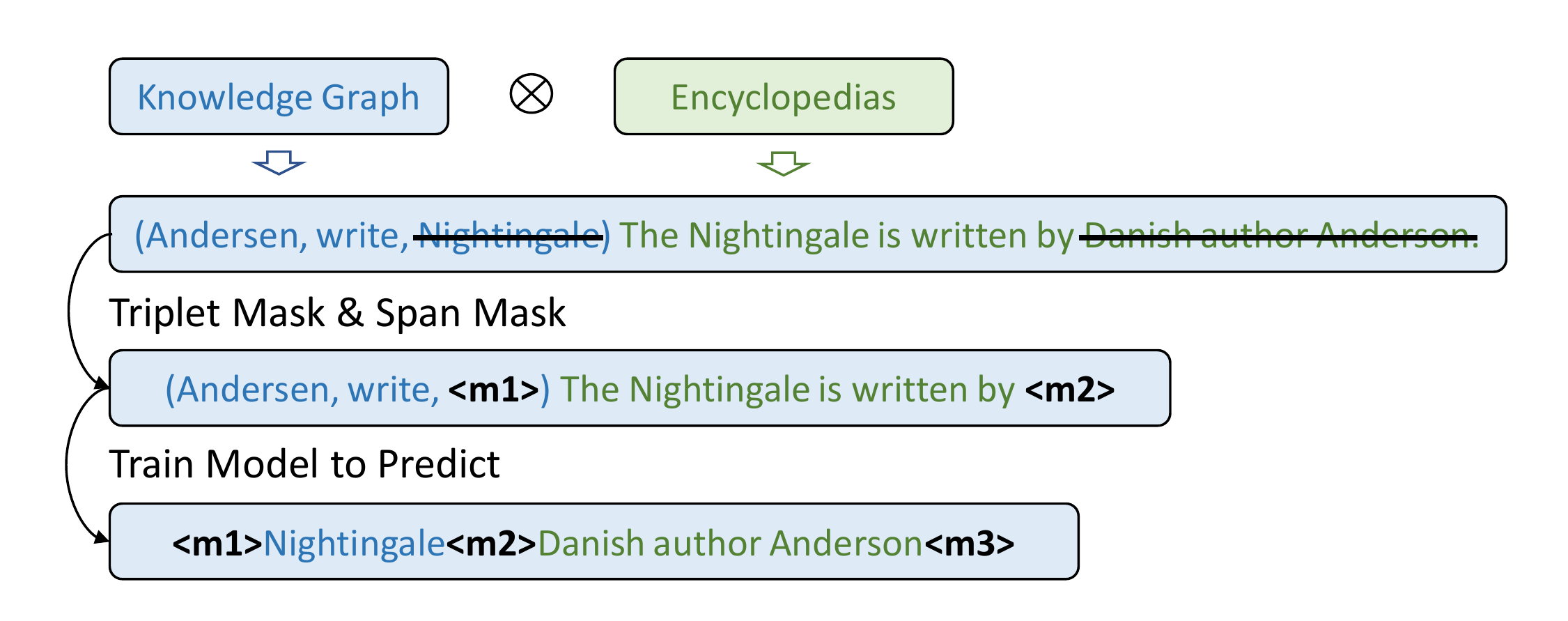}
    \caption{Knowledge enhancement pre-training method based on triple masking (KETM)
    }
    \label{fig:ketm}
\end{figure*}

\section{The Large PLM: BBT-FinT5}
To enhance the performance of the Chinese financial NLP baseline and foster the growth of the open-source community in this domain, we introduce the FinT5 model. This model's architecture and pre-training tasks are consistent with the T5 \cite{2019t5} model and are pre-trained on BBT-FinCorpus (refer to Section~\ref{sec:fincorpus}). We chose this model for its robust performance on many general benchmarks and compatibility with understanding and generating tasks based on the text-to-text paradigm, which facilitates transfer learning. Our experiments demonstrate that the FinT5 model significantly outperforms T5 trained on the general corpus.

In this section, we first describe the architecture and pre-training task of the T5 model. Then we outline the pre-training acceleration method based on DeepSpeed, and finally introduce the knowledge enhancement pre-training method that we propose for the T5 model, which is based on triple masking.

\subsection{Pre-training Model Architecture and Task}
~\citet{2019t5} model all NLP tasks in a text-to-text format which enable the use of a unified network architecture, training approach, and loss function to handle all NLP tasks, promoting transfer learning in the NLP field. Building upon this, they conducted a series of comparative experiments and chose to develop a large-scale PLM, T5, based on an encoder-decoder architecture and pre-trained using MLM. Specifically, T5 utilizes the span mask method proposed by SpanBERT~\cite{joshi2020spanbert}, randomly masking 15\% contiguous spans within a sentence rather than independent tokens.

\subsection{Pre-training Acceleration}

We use the optimizer state parallelism and gradient parallelism implemented by DeepSpeed~\cite{rasley2020deepspeed} to accelerate the pre-training process. In particular, we found that using the BFLOAT16~\cite{kalamkar2019study} half-precision floating-point format for optimization can effectively solve the problem of gradient overflow that occurs in the training process with FP16 half-precision floating-point format, without the need to repeatedly adjust gradient scaling coefficients and other hyperparameters. ~\citet{kalamkar2019study} pointed out that in the training of deep neural networks, the value range (i.e., exponent range) of the floating-point numbers used to represent each parameter in the network is more important for training stability and performance than their mantissa precision. Therefore, the BFLOAT16 format uses the same eight-bit exponent as the FP32 format to represent the same exponent range as the FP32 format, at the cost of having three fewer mantissa bits than the FP16 format. Extensive experiments have shown that this trade-off makes the BFLOAT16 format as fast and memory-efficient as the FP16 format while having training stability and performance close to that of the FP32 format.

\subsection{Knowledge Enhancement Pre-training Method Based on Triple Masking}
We propose a knowledge enhancement pre-training method based on triple masking (KETM).

First, for each triple in the knowledge graph, we use the distant supervision algorithm to obtain sentences corresponding to it. Specifically, for a knowledge triple (head entity, relation, tail entity), if there is a sentence in the encyclopedia that contains both the head and tail entities, we consider this sentence to contain the knowledge described by this triple.

Next, for a sentence and its contained triple, we concatenate the triple at the beginning of the sentence. For the triple part, we randomly mask one element, and for the sentence part, we randomly mask 15\% of a random-length span. Finally, we input the masked triple and sentence into the model and require the model to predict the masked element, as shown in the Figure~\ref{fig:ketm}. The model is trained to fill the masked element in the triple based on the two unmasked elements in the triple and the partially masked sentence, which helps the model better understand and memorize entity-related knowledge.

\section{The Benchmark: BBT-CFLEB}
In this section, we first describe the method used for selecting tasks for the benchmark. We then introduce the selected tasks and the three leaderboards, each of which is composed of different tasks.

\subsection{Task Selection}
We propose that for domain-specific NLP evaluation benchmarks, special attention should be paid to their practicality, especially for the financially valuable field, to better reflect the model's ability in practice. Therefore, we use a practicality score to measure the practicality of the tasks we collect. Specifically, we invited financial experts to evaluate the practicality of each task and gave a low, medium, or high practicality rating, only selecting tasks with a high practicality rating as candidate tasks. In addition, we only kept tasks with a clear open-source statement as candidate tasks. Finally, we selected six tasks for BBT-CFLEB in Table~\ref{tab:fin-datasets}.

\subsection{Task Introduction}

\begin{table*}[!htb]
\centering
\begin{tabular}{l p{6cm} c c c }
\hline
\textbf{Task Name} & \textbf{Introduction} & \textbf{Data} & \textbf{Evaluation} \\ \hline
FinNL & Multi-label classification of financial news & 8000/1000/1000 & F1-score \\ 
FinNA & Generation of summaries for financial news & 24000/3000/3000 & Rouge \\ 
FinRE & Entity relation classification for financial news & 7454/1489/3727 & F1-score \\ 
FinFE & Sentiment classification of financial social media text & 8000/1000/1000 & Accuracy \\
FinQA & Question-answering for financial news/events & 16000/2000/2000 & F1-score \\ 
FinNSP & Detection of negative messages and entities in financial news & 4800/600/600 & F1-score \\ \hline
\end{tabular}
\caption{Summary of CFLEB tasks.}
\label{tab:CFLEB-tasks}
\end{table*}

CFLEB includes six tasks in total, consisting of two language generation tasks and four language understanding tasks. These tasks are as follows:
\begin{itemize}
\item FinNL, a financial news classification dataset. Given financial news articles, the model needs to classify them into up to 15 possible categories, with evaluation measured by F1-Score. The training set contains 8,000 articles, the validation set contains 1,000 articles, and the test set contains 1,000 articles.
\item FinNA, a financial news summarization dataset. Given financial news articles, the model needs to generate a summary, with evaluation measured by Rouge~\cite{lin2004rouge}. The training set contains 24,000 articles, the validation set contains 3,000 articles, and the test set contains 3,000 articles.
\item FinRE, a financial news relation extraction dataset. Given financial news articles and head-tail entity pairs, the model needs to classify the relation between entity pairs into up to 44 categories, including the null relation, with evaluation measured by F1-Score. The training set contains 7,454 articles, the validation set contains 1,489 articles, and the test set contains 3,727 articles.
\item FinFE, a financial social media text sentiment classification dataset. Given financial social media text, the model needs to classify the sentiment of the text into negative-neutral-positive categories, with evaluation measured by accuracy. The training set contains 8,000 articles, the validation set contains 1,000 articles, and the test set contains 1,000 articles.
\item FinQA, a financial news announcement event question-answering dataset, derived from the DuEE-fin~\cite{han2022duee} dataset. Given financial news or announcement text and a question related to an event mentioned in the text, the model needs to generate an answer to the question based on the text, with evaluation measured by F1-Score. The training set contains 16,000 articles, the validation set contains 2,000 articles, and the test set contains 2,000 articles.
\item FinNSP, a financial negative news and its subject determination dataset. Given financial news or social media text and entities mentioned in the text, the model needs to determine if the text contains negative news related to any entity and identify which entity is the subject of the negative news, with evaluation measured by F1-Score. The training set contains 4,800 articles, the validation set contains 600 articles, and the test set contains 600 articles.
\end{itemize}

\subsection{Leaderboard Introduction}
We have organized the tasks into multiple leaderboards according to different ability requirements~\cite{xu2020clue}, so that researchers can observe the model's ability rankings from different perspectives. The leaderboards of FinCUGE are as follows:
\begin{itemize}
\item Overall leaderboard: includes all six tasks.
\item Understanding ability leaderboard: includes four language comprehension tasks, FinNL, FinRE, FinFE, and FinNSP.
\item Generation ability leaderboard: includes two language generation tasks, FinNA and FinQA.
\end{itemize}

\section{Experiments}

In this section, we first introduces the basic settings of the experiment, including the basic information of the PLMs involved in the comparison and the processing format of the tasks in the evaluation benchmark. Then we conduct sufficient experimental and comparative analysis to validate the effectiveness of the proposed model and method. 

\begin{table*}[!htb]
\centering
\resizebox{2.1\columnwidth}{!}{
\begin{tabular}{l c c c c|c|c c|c|c}
\hline
\textbf{PLMs} & \textbf{FinFE} & \textbf{FinNL} & \textbf{FinNSP} & \textbf{FinRE} & \textbf{Un.Avg.} & \textbf{FinNA} & \textbf{FinQA} & \textbf{Ge.Avg.} & \textbf{Avg.} \\
\hline
GPT2-base & 79.05 & 84.09 & 91.30 & 36.37 & 72.70 & 44.19 & 75.22 & 59.71 & 68.37 \\
T5-base & 79.40 & 87.48 & \textbf{95.43} & 54.93 & 79.56 & 48.54 & 83.58 & 66.06 & 74.89 \\
FinBERT-base & 79.45 & 84.69 & 69.01 & 55.33 & 72.37 & - & - & - & -\\
Mengzi-BERT-base-fin & 79.50 & 85.88 & 71.72 & 58.25 & 73.59 & - & - & - & -\\
BBT-FinT5-base & 80.19 & 87.55 & 94.50 & 60.62 & 80.21 & 50.06 & 84.82 & 67.44 & 76.29 \\
BBT-FinT5-base-KE & 79.43 & 87.77 & 95.05 & 61.79 & 80.26 & 51.36 & 85.66 & 68.51 & 76.84 \\
BBT-FinT5-large & \textbf{80.24} & \textbf{88.44} & 94.54 & \textbf{61.88} & \textbf{81.78} & \textbf{51.42} & \textbf{85.95} & \textbf{68.69} & \textbf{77.07} \\
\hline
\end{tabular}
}
\caption{Results of BBT-CFLEB from different PLMs.}
\label{table:fint5}
\end{table*}

\subsection{Experiments Setup}

\subsubsection{Pre-trained Language Models}

The models participating in the comparative experiment of this section include:
\begin{itemize}
\item \textbf{GPT2-base}~\cite{zhao2019uer}. \quad A Chinese GPT2 released by~\citet{zhao2019uer}. Pre-trained using the general corpus CLUECorpusSmall~\cite{xu2020clue}.
\item \textbf{T5-base}~\cite{zhao2019uer}. \quad A Chinese T5 released by~\citet{zhao2019uer}. Pre-trained using the general corpus CLUECorpusSmall~\cite{xu2020clue}.
\item \textbf{FinBERT}~\cite{FinBERT}. \quad A Chinese BERT for the financial domain released by~\citet{FinBERT}.
\item \textbf{Mengzi-BERT-base-fin}~\cite{zhang2021mengzi}. \quad A Chinese BERT for the financial domain released by~\citet{zhang2021mengzi}.
\item \textbf{FinT5-base}. \quad Our Chinese pre-trained language model for the financial domain, pre-trained on our financial corpus, FinCorpus. Its model architecture, parameter size, and pre-training hyperparameters are the same as T5-v1.1-base.
\item \textbf{FinT5-base-KE}. \quad Knowledge-enhanced version of FinT5-base, enhanced by KETM method using CN-DBPedia~\cite{xu2017cn} knowledge graph.
\item \textbf{FinT5-large}. \quad Our proposed Chinese pre-trained language model for the financial domain, with a total of about 1 billion model parameters, and the pre-training hyperparameters are the same as T5-base.
\end{itemize}

\subsubsection{Fine-tuning}
For generative models (GPT, T5), we evaluated all six datasets by modeling all tasks as text-to-text. For BERT-based models, we evaluated them on four language understanding tasks: FinNL, FinRE, FinFE, and FinNSP, using BERT with an additional classification layer for all tasks.

\subsection{Experiment 1: Comparison of Pre-trained Model Architectures}
For the two models in the general domain, GPT2-base and T5-base, their pre-training corpora, hyperparameters, and training volume are all the same, but their average scores differ significantly, with T5-base significantly outperforming GPT2-base, as shown in Table~\ref{table:fint5}. This difference is mainly due to the differences in the architectures, parameter sizes, and pre-training methods of the T5 and GPT models. This performance confirms the correctness of our choice of the T5 model.

\subsection{Experiment 2: Effectiveness of Domain Pre-training}
As shown in Table~\ref{table:fint5}, the comparison between the FinT5-base model and the T5-base model indicates that the FinT5-base model pre-trained on FinCorpus significantly outperforms the T5-base model with the same parameter size, demonstrating the effectiveness of domain pre-training and the effectiveness of FinCorpus.

\subsection{Experiment 3: Superiority Compared to Existing Models in the domain}
As shown in Table~\ref{table:fint5}, in the four language understanding tasks evaluated with FinBERT and Mengzi-BERT-base-fin, FinT5-base significantly outperformed both models, demonstrating the superiority of FinT5 over existing models in the domain.

\subsection{Experiment 4: Effectiveness of KETM}
As shown in Table~\ref{table:fint5}, by comparing FinT5-base-ke with FinT5-base, it can be seen that the knowledge-enhanced text modeling method significantly improves the model's performance on tasks such as relation extraction and news summarization, without significantly compromising the performance on other tasks, thus proving the effectiveness of the KETM method.

\subsection{Experiment 5: Effectiveness of parameter scaling up}
As shown in Table~\ref{table:fint5}, the performance comparison between FinT5-base and FinT5-large models indicates that the FinT5-large model with one billion parameters performs significantly better than the FinT5-base model, demonstrating the effectiveness of parameter scaling up.

\section{Conclusion}

In this article, we introduced three new contributions to the domain of NLP in the context of Chinese finance. We created the largest open-source corpus for this domain, called FinCorpus, which contains a diverse collection of around 300GB of text from four sources. Our FinT5 model is the largest pre-trained language model for the Chinese financial domain, with one billion parameters. To enhance our pre-training method, we developed a unique knowledge-based approach called KETM, which was effective. We also created a benchmark to evaluate the understanding and generation capabilities of language models, called CFLEB. We believe domain benchmarks should prioritize practicality to better reflect how improvements in language models in academia can benefit the real world. Our future work includes expanding FinCorpus and FinT5 and exploring multilingual and multimodal applications.

\newpage

% Entries for the entire Anthology, followed by custom entries
\bibliography{anthology}
% \bibliography{custom}
\bibliographystyle{acl_natbib}

% \appendix

% \section{Example Appendix}
% \label{sec:appendix}

% This is a section in the appendix.

\end{document}